\documentclass[conference]{IEEEtran}
\IEEEoverridecommandlockouts
\usepackage{cite}
\usepackage{amsmath,amssymb,amsfonts}
\usepackage{algorithmic}
\usepackage{graphicx}
\usepackage{textcomp}
\usepackage{xcolor}
\usepackage[linesnumbered,lined,boxed,commentsnumbered,ruled,longend]{algorithm2e}               
\usepackage{url}            
\usepackage{booktabs}       
\usepackage{multirow}       
\usepackage{subfigure}      
\makeindex

\SetKwInput{KwData}{Input}
\SetKwInput{KwResult}{Output}
\SetKwFor{ParFor}{for}{do in parallel}{end for}

\def\BibTeX{{\rm B\kern-.05em{\sc i\kern-.025em b}\kern-.08em
    T\kern-.1667em\lower.7ex\hbox{E}\kern-.125emX}}

\begin{document}

\title{Personalized Decentralized Federated Learning\\ with Knowledge Distillation\\
\thanks{The work of E. Jeong is supported by the Huawei-EURECOM Chair on Future Wireless Networks.}
}

\author{
    \IEEEauthorblockN{Eunjeong Jeong and Marios Kountouris}
    \IEEEauthorblockA{Communication Systems Department\\
    EURECOM\\
    Sophia Antipolis 06410, France
    \\\{eunjeong.jeong, marios.kountouris\}@eurecom.fr}
}

\maketitle

\begin{abstract}
  Personalization in federated learning (FL) functions as a coordinator for clients with high variance in data or behavior. Ensuring the convergence of these clients' models relies on how closely users collaborate with those with similar patterns or preferences. However, it is generally challenging to quantify similarity under limited knowledge about other users' models given to users in a decentralized network. To cope with this issue, we propose a personalized and fully decentralized FL algorithm, leveraging knowledge distillation techniques to empower each device so as to discern statistical distances between local models. Each client device can enhance its performance without sharing local data by estimating the similarity between two intermediate outputs from feeding local samples as in knowledge distillation. Our empirical studies demonstrate that the proposed algorithm improves the test accuracy of clients in fewer iterations under highly non-independent and identically distributed (non-i.i.d.) data distributions and is beneficial to agents with small datasets, even without the need for a central server. 
\end{abstract}

\begin{IEEEkeywords}
decentralized federated learning, personalization, knowledge distillation
\end{IEEEkeywords}

\section{Introduction}
Since the appearance of federated learning (FL)~\cite{konevcny16-FL} as a promising and efficient solution for distributed learning with collaborative clients, numerous research studies have investigated this paradigm in distributed networks of users under various hindrance factors, such as limited local storage~\cite{imteaj21}, information leakage~\cite{geiping20}, biases across user experiences~\cite{sattler19}, and transmission impairments~\cite{niknam20}.
The main objective of FL and of many of its decentralized variants is generally to acquire a global model across all devices. Nevertheless, a single common model deduced from all participants may not satisfy the clients whose tasks or data distributions significantly deviate from the rest. On this account, personalized FL~\cite{marfoq21-localmem, huang20, fallah20-perfedavg, wu22-motley, mansour20-hypcluster, jiang19-fedavg+, divi21} has been considered as a means to provide a customized solution to users with statistical heterogeneity.
A widely used procedure for personalized FL first constructs a global model using a central aggregator as a draft, and then customizes it under each agent's control.~\cite{mansour20-hypcluster, jiang19-fedavg+, divi21}
On the other hand, personalized FL is particularly commensurate with serverless networks as each agent executes the training process autonomously, thereby enabling asynchronous learning~\cite{vanhaesebrouck17, zantedeschi20, ye22, li22_L2C-cvpr, borodich21, sadiev22, bellet18}. 

Despite its practical relevance, there are several issues and technical challenges associated to personalized federated learning, which we briefly discuss below.
First, among users with diverse characteristics, each agent has to find sufficiently similar peers since putting more weights on received model parameters with higher similarity during superposition improves its performance~\cite{bellet18}.
Unfortunately, choosing the right distance measurement, which can capture the actual similarity, is challenging. Moreover, the system has to encounter a tradeoff between model complexity and metric complexity. If the exchanged information has a more complex structure, participants have no choice but to use simpler metrics~\cite{marfoq21-neurips}.
Second, FL fundamentally requires ensuring privacy within each agent, which commonly implies but is not limited to maintaining data samples private. For instance, the information on possessed class labels for classification tasks should also be kept private~\cite{jeong18}. However, because of their omniscient viewpoint, most existing personalized and decentralized learning schemes do not thoroughly preserve inter-device privacy. Occasionally, the agents take a constant for local training, which is derived from public knowledge, e.g., the number of others' training samples~\cite{vanhaesebrouck17}. Even though the variables used for training are perfectly separated, a decentralized collaborative learning scheme introduced in \cite{ye22} assumes the presence of a proxy dataset accessible to anyone. In~\cite{marfoq21-localmem, zantedeschi20}, the selected agents request to exchange hypotheses represented as a weighted sum of base data distributions. These identify the direct information of which class labels one possesses from its neighbors.

To address the aforementioned challenges, we propose KD-PDFL, a personalized decentralized FL scheme that provides a completely enclosed service. With KD-PDFL, each client can individually update all parameters from their first-person perspective, including the connectivity graph, the local model, and the local dataset. In this approach, each user receives only model copies from its neighbors and determines their optimal combination.
This property differentiates our algorithm from prior works, which either ask users to seek additional external information or rely on isolated personalization methods like local fine-tuning. 

As a powerful tool that enhances inference capability of clients with simple models, we introduce knowledge distillation into our proposed scheme, where agents evaluate similarity with co-distillation based on local validation datasets. Thanks to embracing the characteristic of distillation, agents are also free from the need for model homogeneity since distillation enables cooperation across models with different layer structures as long as the models have a common layer with the same dimension.
Our experimental results show that KD-PDFL achieves higher test accuracy within smaller global iterations compared to other personalized decentralized FL schemes, given that the amount of exchanged information is the same. We also provide a guideline for tuning the hyperparameters used in implementing our experiments.

\section{Preliminaries}\label{subsect:prelim}

In this section, prior to introducing our work, we provide a brief overview of the two major ingredients of our proposed solution: (i) personalized decentralized learning, which describes the overall protocol of how users exchange information; (ii) knowledge distillation, which elaborates on how they draw relevance from others.

\subsection{Personalized decentralized learning}
\begin{figure}
    \centering
    \includegraphics[width=\columnwidth]{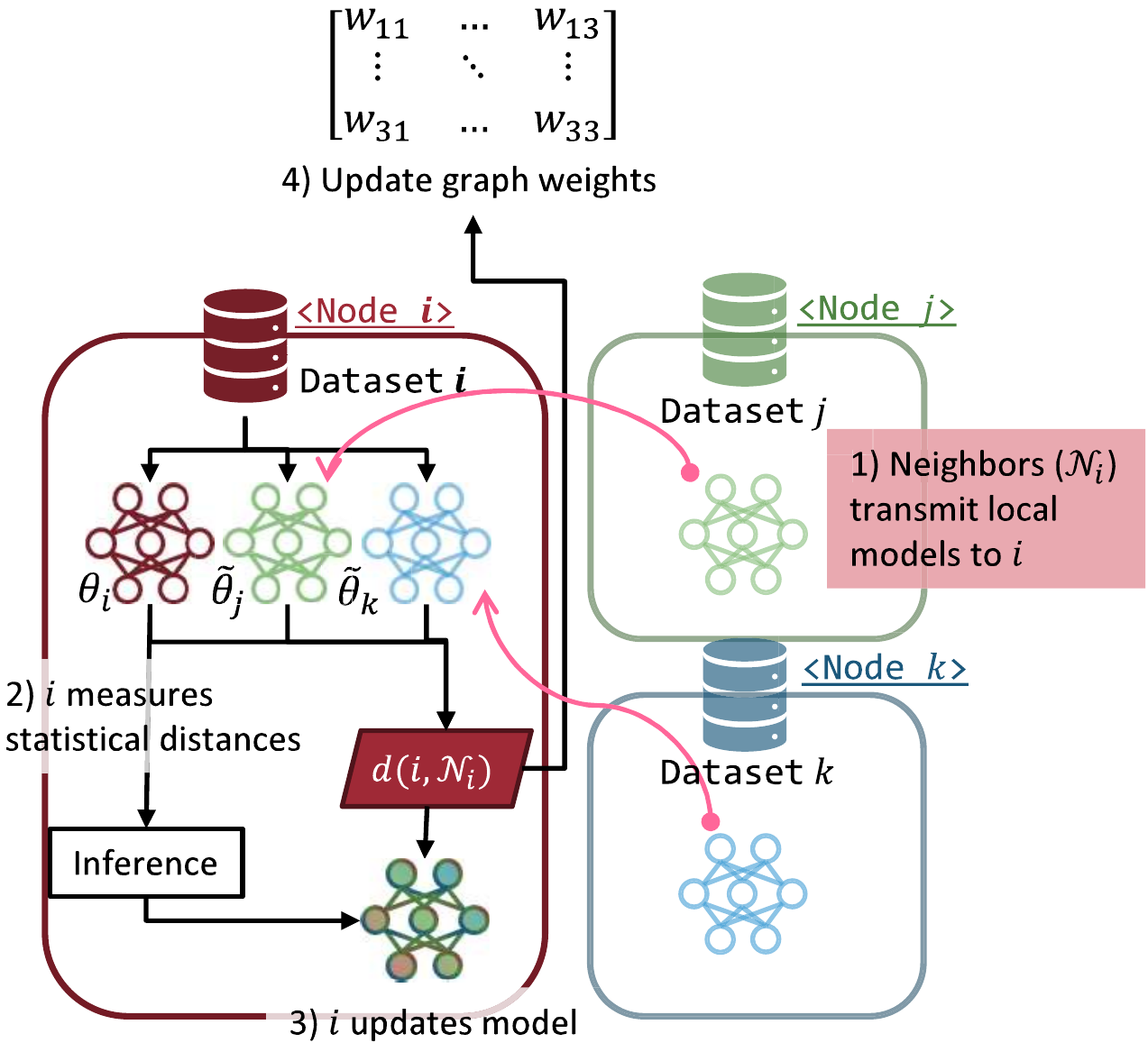}
    \caption{A general schematic of personalized decentralized federated learning.}
    \label{fig1}
\end{figure}

In contrast to distributed networks with a central server, a fully decentralized network implies that no node has authority or accessibility to construct a global consensus model at any instant of learning process. Thus, personalized and decentralized FL naturally exclude building a common model before local customization.
We consider a decentralized learning system that assigns different central entities for each global iteration. In this network, a randomly selected node wakes up to serve as a temporary center node, which is also called a ``star node''. First, this star node collects gradient parameters from its neighbors. Subsequently, it calculates inferences from all received gradients, which are used to measure the similarities among the learning objectives thereafter. Users manage the traits of these similarities by selecting and focusing on communicating with neighbors that have the highest similarities. In literature, these semantic traits are modeled using a \emph{collaboration graph}, which can be a property of communication links between nodes or a target variable to optimize mixing weights. A collaboration graph is a weighted graph whose edge weights represent quantified relevance between learning tasks. (See also Fig.~\ref{fig1})

Once the star node finishes the calculation described above, it updates the weights of the collaboration graph according to the similarity variables. Meanwhile, it also updates its model using a weighted sum of all received parameters, where the weights are those obtained from the collaboration graph.

\subsection{Knowledge Distillation (KD) and co-distillation (CD)} KD~\cite{hinton15-kd} is a knowledge transfer method in which multiple clients can compare the outputs from a common dataset to absorb the inferred knowledge of the others. They quantify the similarity between the logits. In CD~\cite{anil18-cd}, all clients are in an equal position to learn from the other as everyone works as a student. KD and CD have the advantage of allowing the users to transfer knowledge between models with heterogeneous structures. Yet, the system must have a public/common dataset in order to let the participants compare each logit one by one with identical batches. This conflicts with the vanilla FL approach that guarantees the data maintenance attribute without transmission of data samples.

\section{Problem Setting}\label{sect:probset}
We consider a joint minimization problem across $M$ users. Each user indicated as $i\in[M]=\{1,2,\cdots,M\}$ has access to a training set $X_i$ and its goal is to minimize its loss function $L_i:\mathbb{R}^{n_\mathcal{X}}\rightarrow \mathbb{R}^+$ where $n_\mathcal{X}$ is the dimension of the input features. The objective of the whole system is to minimize the total local loss and the jointly calculated dissimilarity. 

We adapt the idea of a collaboration graph $W$ to represent the connectivity between any two nodes in the network. A collaboration graph $W = [w_1, w_2, \dots, w_M]$ is a matrix of stacked connectivity vectors of each client. Each column vector of $W$, denoted as $w_i=[w_{i1},\dots,w_{iM}]^T$, is a connectivity vector of client $i$ whose elements are the edge weights. In this work, the term ``connectivity'' implies relevance across two nodes, thus $w_{ij}\geq0$ is proportional to the degree of recognition of node $i$ for how relevant node $j$'s task is to its own task.

\begin{figure*}[!t]
    \centering
    \includegraphics[width=0.85\textwidth]{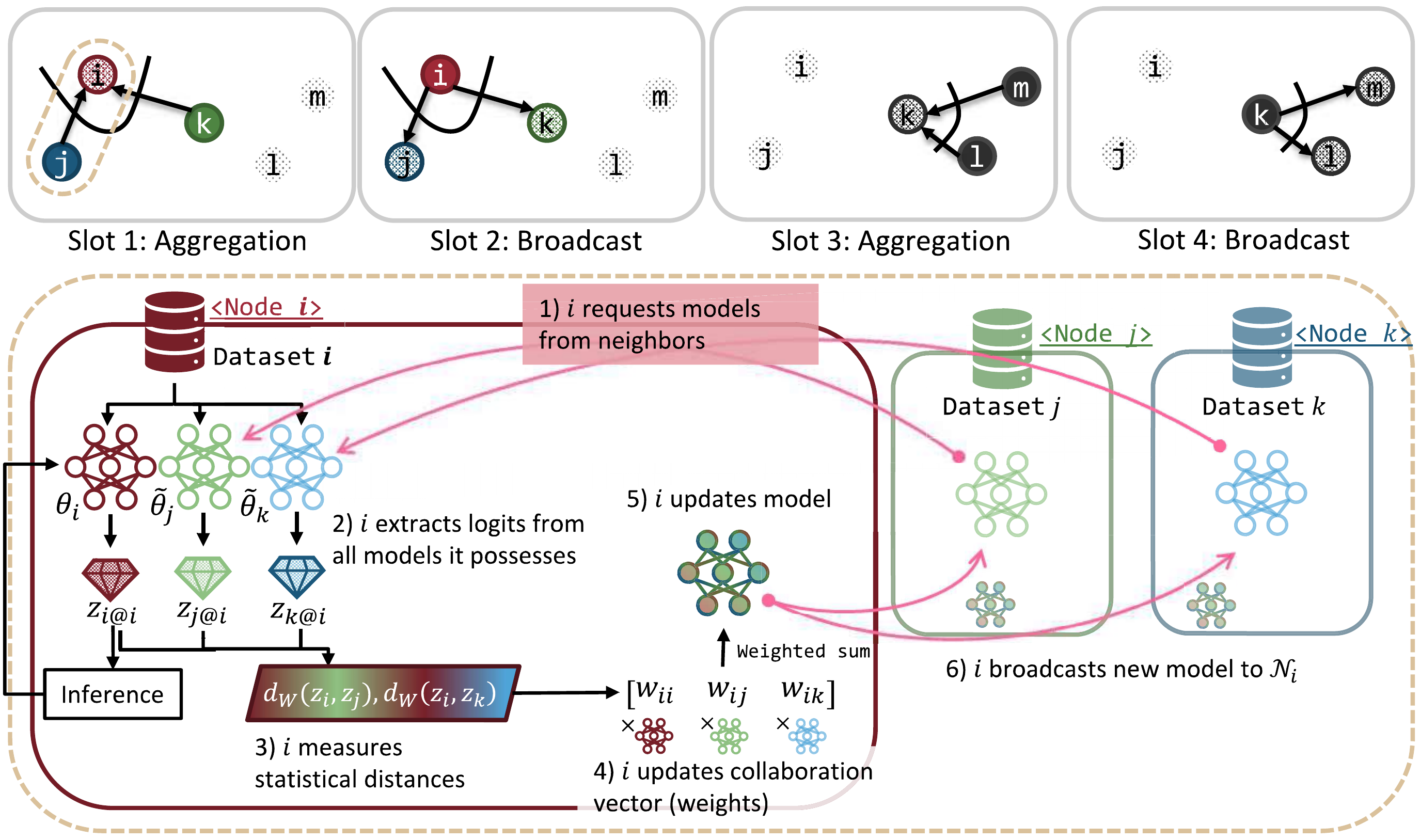}
    \caption{Above: communication protocol during exchange intervals. Below: model and collaboration vector update elaborated. The figure illustrates the process executed only in the star node.}
    \label{figmain}
\end{figure*}

\textbf{Objective function:}
The main target of our system is to learn the personalized models $\Theta=\{ \theta_1,...,\theta_M \}$ and the collaboration graph $W\in\mathbb{R}^{M^2}$ that minimize the following joint optimization problem
\begin{multline}\label{eq:obj-fn}
    \min_{\Theta, W} J(\Theta, W) = \sum_{i=1}^{M} L_i\left(\theta_i; X_i\right) \\
    + \frac{\mu_1}{2}\sum_{i,j\in[M]} w_{ij} d(i,j)  + \mu_2 g(w)
\end{multline}
where $L_i(\cdot;\cdot)$ is a local loss function and $d(i,j)$ is the measured distance (dissimilarity) between the model estimations of two clients $i$ and $j$. The second term allows assessing task relevance by penalizing the links between any two nodes with large statistical distances. The third term $g(w)$ is a regularization term that strongly encourages the users to participate in collaboration by giving a high penalty when a user tries only local training. $\mu_1$ and $\mu_2$ are hyperparameters for adjusting the influence of each term above, respectively.

\section{Personalized Decentralized Learning with Knowledge Distillation}\label{sect:KD-PDFL}
We consider a fully decentralized network where agents do not implement strict synchronization and cooperate through peer-to-peer communication. Particularly, our definition of decentralization imparts autonomy in determining the collaboration graph. In other words, each user evaluates the collaborative weights independently and privately instead of accessing a row of a connectivity matrix shared in public. 
The clients follow a cross-silo setting where each client performs all steps of the learning process, i.e., has datasets locally distributed for training, validation, and test purposes. 
In this section, we focus on the method to extract relevant information across the nodes in a serverless fashion. In every exchange interval $T_{ex}$, the agents follow the step-by-step instructions below (see also Fig.~\ref{figmain}):
\begin{enumerate}
    \item A node assigned as a star node, say user $i$, wakes up to ask for transmission from a group of its neighbors at time $t$, $\mathcal{N}_i^{(t)}$. The peers send their local model updates to the star node $i$. At the end of the transmission, $i$ has $|\mathcal{N}_i^{(t)}|$-copies of model parameters of its neighbors if no packet loss has occurred.
    \item From $i$'s local training dataset $X_i$, $i$ packs training samples in a batch $B_i$. The batch $B_i$ is fed to model parameters of all $j\in\mathcal{N}_i^{(t)}$ in order to get intermediate outputs (e.g., logits), denoted as $z_{ij}$.
    \item Node $i$ measures the statistical distance $d_W(i,j)$ for all $j\in\mathcal{N}_i^{(t)}$, then updates $w_i$ of which the gradient function $\nabla h_i(w)$ is the partial derivative of eq.~(\ref{eq:obj-fn}) with respect to the computation entity $i$. Based on this gradient, node $i$ updates a collaboration weight of $j$ from the viewpoint of $i$, which is always bounded to a nonnegative value. The following equation illustrates the policy of updating the connectivity vector of $i$:
        \begin{equation}
        \begin{aligned}
            & \nabla h_i(w_i^{(t)}) = \mu_1 d_{W, i} + \mu_2 \nabla g_i(w_i^{(t)}) \\
            & \eta^{(t)} = 1./ |\nabla h_i(w_i^{(t)})|  \\
            & w_{ij}^{(t+1)}= \max (0, w_{ij}^{(t)}-\eta^{(t)} \nabla h_i(w_i^{(t)}))
            \label{eq:connvectorupdate}
        \end{aligned} 
        \end{equation}
        where $d_{W, i} = [d_W(i,1),\cdots,d_W(i,M)]$ indicates the statistical distance vector of node $i$.
    \item $i$ updates its connectivity vector using $d_W(i,j)$.
    To quantify the distance between two logits that are formed as probability distributions, the given user calculates the batched mean of the Wasserstein distance between the two variables. In particular, under classification tasks where the probability distribution is discrete and the number of possible classes is known, a user can compute the arithmetic mean of all Wasserstein distances from each single data sample of the batch. With $n_i$ local batch samples and $n_L$ possible classification labels, the Wasserstein distance of two logits with $(n_i \times n_L)$ size is computed as follows:
        \begin{align}
            d_W(i,j) = \frac{1}{n_i} \sum_{x=1}^{n_i} \sum_{l=1}^{n_L} \| p_{i,l}^{(x)} - p_{j,l}^{(x)} \|_2^2
            \label{eq:wasserstein2d}
        \end{align}
    where $p_{i,l}^{(x)}$ indicates the logit for class label $l$ that goes through user $i$'s model from an input data point with index $x$.
    Note that the importance of $i$ from the viewpoint of $j$ may differ from that of $j$ from the viewpoint of $i$ due to differences in model complexity or the number of trainable samples. Thus, $W$ is not symmetric, and agents do not need to share updated connectivity weights.
\item A new model for $i$ is a weighted sum of all footprints it has at that time, where the weights are the elements of the connectivity vector.
    \item In the broadcast phase at the next time slot, node $i$ broadcasts $\theta_i^{(t)}$ to the peer group of the previous time slot, $\mathcal{N}_i^{(t-1)}$.
    \item Other than the exchanging iterations, each agent performs only local learning.
\end{enumerate}

One cannot find its own weight $w_{ii}$ of the connectivity vector since having statistical distance with itself does not make sense. For that, we adopt a widely used concept named confidence to indicate the weight of local model updates. A confidence of node $i$ at time $t$, denoted as $c_i$, is defined as a time-varying term dependent on the size of its local training set:
    \begin{align}
        c_i^{(t)} = \min \left( \frac{|X_i|}{c_{base}},\ \frac{1}{|\mathcal{N}_i^{(t)}|+1}\right)
        \label{eq:confidenceconstant}
    \end{align}
where $c_{base}$ is a constant for a base confidence that is neither induced from shared information across the devices nor to be shared with others. The algorithmic description of KD-PDFL can be found in Appendix~\ref{subsect:alg}.

\section{Experiments}\label{sect:eval}
\begin{figure}[!t]
    \centering
    \includegraphics[width=\columnwidth]{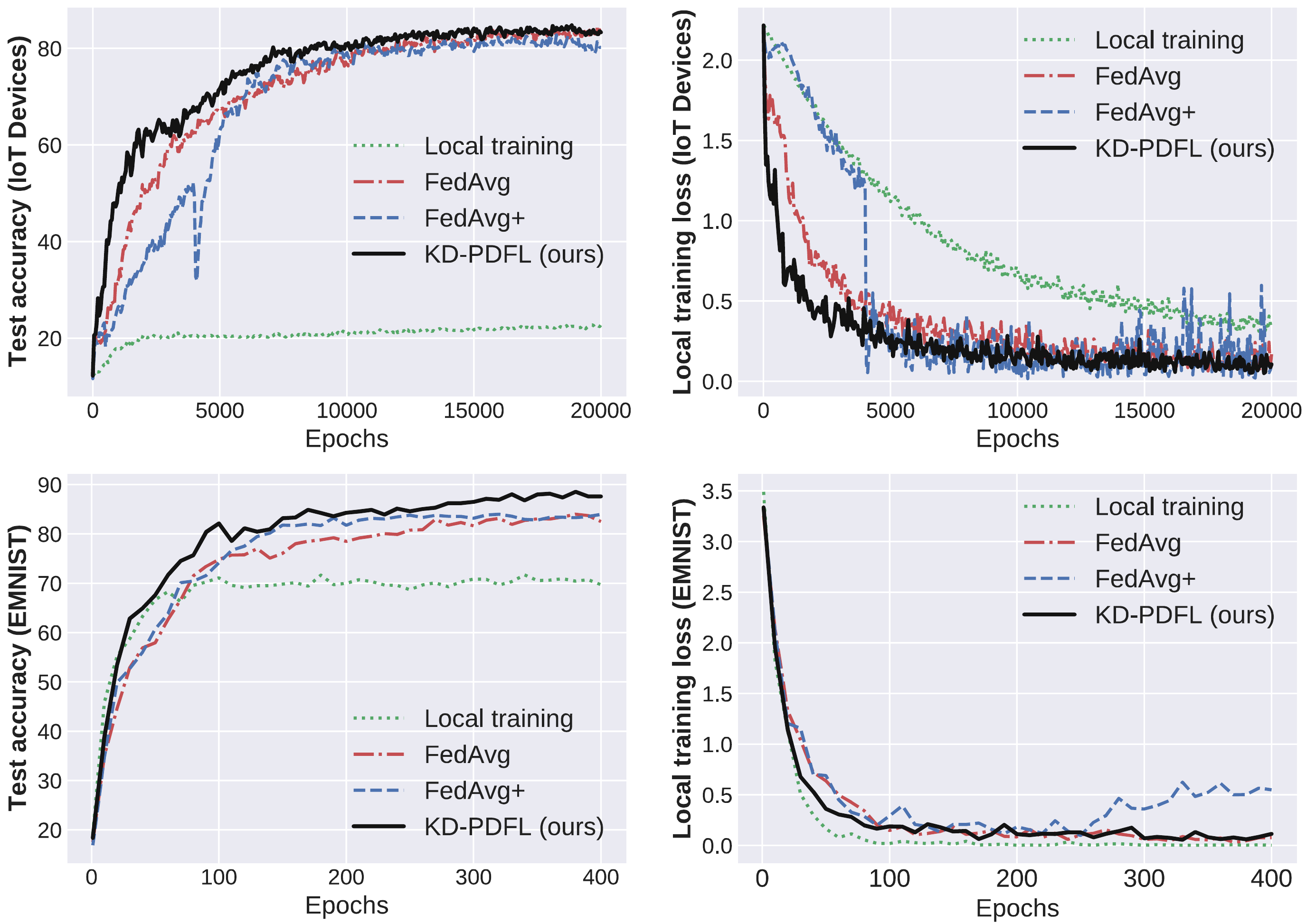}
    \caption{Learning progress of standalone and collaborative users in different collaboration methods over iterations ($M=40$).}
    \label{figaccloss}
\end{figure}

In this section, we evaluate the performance of KD-PDFL in terms of per-client test accuracy. We compare the proposed scheme with three related baseline scenarios for decentralized networks: (i) standalone case where all users perform local training only; (ii) conventional federated averaging (FedAvg~\cite{mcmahan17-fedavg}), which returns the arithmetic mean of received model parameters; and (iii) FedAvg followed by fine-tuning, which switches the algorithm into Reptile~\cite{nichol18-reptile}. (FedAvg+~\cite{jiang19-fedavg+})
Note that in our evaluations, FedAvg corresponds to a decentralized learning scheme but with non-personalized service, i.e., it aims to build a single global model for all clients.

\begin{table}[!t]
\centering
\resizebox{\columnwidth}{!}{%
\begin{tabular}{@{}clccc@{}}
\toprule
Dataset                                                                & \multicolumn{1}{c}{Methods} & M=10            & M=20            & M=40            \\ \midrule
\multirow{4}{*}{\begin{tabular}[c]{@{}c@{}}IoT\\ devices\end{tabular}} & Local learning              &                 & 0.210$\pm$0.089 &                 \\
                                                                       & FedAvg                      & 0.759$\pm$0.650 & 0.643$\pm$0.107 & 0.610$\pm$0.493 \\
                                                                       & FedAvg+                     & 0.802$\pm$0.028 & 0.726$\pm$0.630 & 0.697$\pm$0.117 \\
                                                                       & KD-PDFL (ours)              & \textbf{0.816$\pm$0.032} & 0.739$\pm$0.040 & 0.716$\pm$0.101 \\ \midrule
\multirow{4}{*}{EMNIST}                                                & Local learning              &                 & 0.697$\pm$0.209 &                 \\
                                                                       & FedAvg                      & 0.764$\pm$0.119 & 0.784$\pm$0.108 & 0.824$\pm$0.142 \\
                                                                       & FedAvg+                     & 0.771$\pm$0.104 & 0.806$\pm$0.123 & 0.841$\pm$0.136 \\
                                                                       & KD-PDFL (ours)              & 0.787$\pm$0.108 & 0.835$\pm$0.116 & \textbf{0.870$\pm$0.082} \\ \bottomrule
\end{tabular}
}
\vspace{10pt}
\caption{Summary of per-client test accuracy under IoT devices ($T_{ex}=20$) and EMNIST datasets ($T_{ex}=5$). }
\label{table}
\vspace{-10pt}
\end{table}

We carry out the experiments with two types of datasets. The first one we consider is a set of smart users practicing classification tasks using web traffic information from IoT devices. The data samples have 296 input features and the output dimension of the dataset is $9$.\footnote{\url{https://www.kaggle.com/datasets/fanbyprinciple/iot-device-identification}} Each user contains $15$ to $100$ training samples with non-i.i.d. distribution and $100$ local test samples, which follows the same distribution with the local training set. We split the IoT devices training dataset into data points of which the class labels follow a symmetric Dirichlet distribution with parameter $0.1$. This data split is the same as in a benchmark in~\cite{wang20} but is more biased due to the smaller parameter, allocating one to four class labels per each user with uneven number of data samples per label. A neural network model in every user includes one batch normalization layer, one rectified linear unit (ReLU) layer, and two linear layers. Each model has $39,769$ trainable parameters in total. 
The other set is for classification tasks using the EMNIST dataset, which is composed of $28\times28$ pixel images of handwritten Roman alphabets and letters. In our experiments, a total of $47$ class labels are included under balanced split settings. Each user has a convolutional neural network (CNN) made of two convolutional layers, two max pooling layers, and two fully connected linear layers, which in total has $970,847$ trainable parameters. For both datasets, $10$ to $40$ users participate per experiment. 
The channel gain between each pair of participants follows Rayleigh fading, resulting in $5$ neighbors reachable on average for each exchanging interval.

Fig.~\ref{figaccloss} shows that equal averaging over all participants, as provided in FedAvg, achieves the poorest performance as the environment is significantly non-i.i.d.~\cite{li19-fedavg-noniid}. Meanwhile, FedAvg+ compensates for the overfitting loss from FedAvg. The fine-tuning process began to adjust the hyperparameters from the moment the iteration reached $t=2000$ in case of experiments with IoT devices dataset and $t=150$ in case of EMNIST, which led to temporal deterioration, higher per-client test accuracies and lower training losses at the end.

Table~\ref{table} shows that KD-PDFL enables the clients to extend the upper bound of their estimated test accuracy without collaboration. Notably, users with small local training sets benefit from increased test accuracy from $21.0$\% to $81.6$\% on average. 
These weak users with small training sets also undergo a more challenging similarity decision. Since the lack of training set incurs blurry distance divergence across intermediate outputs that they calculate, estimating connectivity weights becomes more difficult.

\begin{figure}
    \centering
    \includegraphics[width=0.9\columnwidth]{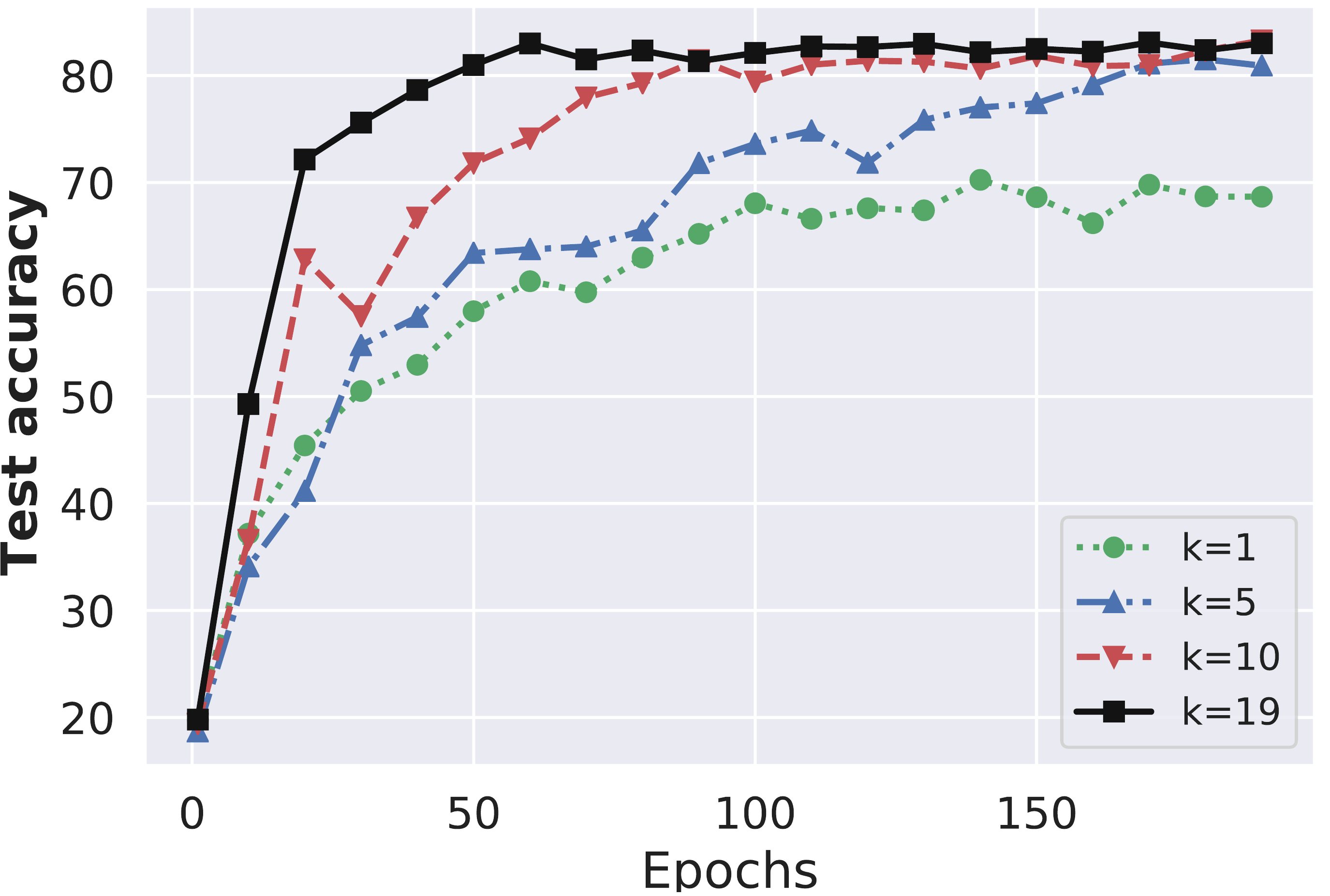}
    \caption{Trend on test accuracy improvement with respect to the number of connecting peers for each communication slot. ($M=20$, $T_{ex}=5$, dataset: EMNIST)}
    \label{figk}
\end{figure}

\begin{figure}
    \centering
    \includegraphics[width=\columnwidth]{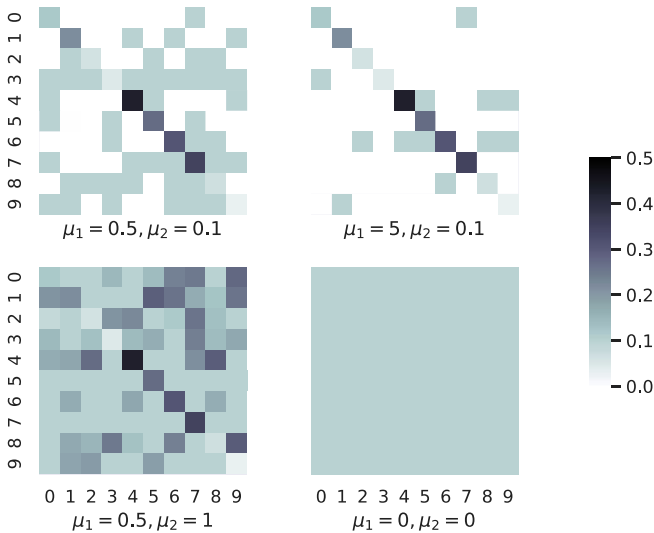}
    \caption{Heatmap of collaboration weight matrix under different values for $\mu_1$ and $\mu_2$.}
    \label{figmu}
\end{figure}

Figuring out the relative importance becomes more manageable when a client can search over a lot of peers at once. Nonetheless, as $M$ increases, settling down on the most effective collaborative graph is more challenging. Fig.~\ref{figk} shows the trend in test accuracy with respect to the number of communicable peers per collaboration. Putting many neighbors into simultaneous consideration ends up walking on the same track with FedAvg, resulting in a prolonged ``getting to know'' session and slower model convergence. On this account, it is recommended to set an upper limit on the number of neighbors accessing per exchange interval, even though the channel conditions are good enough to cover many neighbors at once.

Fig.~\ref{figmu} visualizes the impact of coefficients of penalty and regularization terms on collaboration graphs. $\mu_1$ is a multiplier on the personalization term of the joint loss function that controls the sensitivity to the statistic distances of the system. $\mu_1\simeq 0$ turns off the effect of statistic distances on the training loss. Meanwhile, $\mu_2$ adjusts the regularization to the weight values; thus, $\mu_2\simeq 0$ may let some participants abandon cooperation and run only local training. A situation in which $\mu_1=\mu_2=0$ is equivalent to federated learning that aims to construct a global model. 

\section{Concluding remarks}\label{sect:conc}
We introduced a distillation-based algorithm for personalized federated learning over fully decentralized networks, leveraging the privacy preservation and the convenience of measuring the statistical distance across clients using logits generated from local storage. Our experimental results showed that the proposed KD-PDFL is a promising decentralized approach compared to other personalized FL methods, with each device having full autonomy in computing, including updating a collaboration graph.

In future work, distillation-based personalization can be extended to unsupervised learning tasks since the distance measurement part does not require class labels as long as the local loss function does not include target labels in its metric.
An interesting problem that arises by connecting topologies and connectivity graphs regarding physical distances among edge devices~\cite{bars22-yes-topology} is how to choose a subset of neighbors of a personalized and decentralized FL system in a communication-efficient way. A crossover with multi-task learning~\cite{smith17-mocha} also remains a potential direction of this work.

\appendix

\subsection{Algorithm of KD-PDFL}\label{subsect:alg}
The function \texttt{Midgetter} in Alg.~(\ref{alg-kd-pdfl}) extracts outputs from the intermediate layers of two neural networks. The intermediate layer can be either the output layer or one of the hidden layers depending on whether the agents transmit the entire models or only the base layers. Feeding the identical batch of client $i$'s training samples, $\theta_i$ and $\theta_j$ returns the logits $z_{i@i}$ and $z_{j@i}$, respectively. The expression $a@b$ refers to an output of client $a$'s model under control of client $b$, i.e., the distillation is conducted at user $b$'s device while using $b$'s computation resources without sharing.
\begin{algorithm}[htbp]
\label{alg-kd-pdfl}
    \SetAlgoLined
    \caption{KD-PDFL: Distillation-based personalized decentralized FL}
    \BlankLine
    \KwData{$\theta_i^{(0)} = \mathbf{0} \in \mathbb{R}^{n_i}$, $w_{ii}=0 \ \forall i\in[M]=\{1,2,...,M\}$, $w_{ij}=1/M \ \forall j\in[M]\setminus i$}
    \KwResult{$\Theta^{(T)},\ W^{(T)}$}
    \DontPrintSemicolon
    \For{$t$ in $(0, T]$}{
    \uIf{$t\equiv 0 \pmod {T_{ex}}$}{
    Random user $i$ wakes up \& draw a subset $\mathcal{N}_i^{(t)}$\;
    \ParFor{each neighbor $j\in\mathcal{N}_i^{(t)}$}{
        Receive $\tilde{\theta}_j^{(t-1)}$ from each $j$\;
        $(z_{i@i}, z_{j@i}) = \texttt{MidGetter}(\theta_i, \tilde{\theta}_j, X_i)$
       \tcp*{find intermediate outputs}
       $d_W(i,j) = \texttt{Wasserstein2D}(z_{i@i}, z_{j@i})$
       \tcp*{find statistic distances}       
       }
       \texttt{ConnVectorUpdate}$(i,j)$ as in Equation (\ref{eq:connvectorupdate}) \;
       Update $\theta_i^{(t+1)}= \sum_{j\in [M]\setminus\{i\}} w_{ij}^{(t+1)}\tilde{\theta}_j^{(t-1)}+c_i w_{ii}^{(t+1)}\theta_i^{(t)}$ \;
       }
    \uElseIf{$t\equiv 1 \pmod {T_{ex}}$}{
        \For{each neighbor $j\in\mathcal{N}_i^{(t-1)}$}{
            Receive $\tilde{\theta}_i^{(t-1)}$ from $i$ \;
            Update $\theta_j^{(t+1)}= \tilde{\theta}_i^{(t)}$ \;
            }
        }
    \Else{
        $\theta_i^{(t+1)} = \theta_i^{(t)} - \eta \nabla f_i(\theta_i^{(t)})$
        \tcp*{Local Update}
    }
}
\end{algorithm}

\bibliographystyle{ieeetr}
\bibliography{ref}

\end{document}